%% file: main.tex
\documentclass{article}
\usepackage{spconf}
\usepackage{amsmath,amssymb,amsfonts}
\usepackage[utf8]{inputenc}
\usepackage{mathtools}
\usepackage{multirow}
\usepackage{graphicx}
\usepackage{import}
\usepackage{color}
\usepackage{flushend}
\usepackage{caption}
\usepackage{subcaption}
\usepackage{booktabs}
\usepackage[draft]{hyperref}

\title{CNN for License Plate Motion Deblurring}
\name{Pavel Svoboda, \qquad Michal Hradiš, \qquad Lukáš Maršík, \qquad Pavel Zemčík}
\address{
	Brno University of Technology \\
	Czech Republic \\
	\{isvoboda,ihradis,imarsik,zemcik\}@fit.vutbr.cz
}

\begin{document}

\maketitle


\begin{abstract}
In this work we explore the previously proposed approach of direct blind deconvolution and denoising with convolutional neural networks in a situation where the blur kernels are partially constrained. 
We focus on blurred images from a real-life traffic surveillance system, on which we, for the first time, demonstrate that neural networks trained on artificial data provide superior reconstruction quality on real images compared to traditional blind deconvolution methods.
The training data is easy to obtain by blurring sharp photos from a target system with a very rough approximation of the expected blur kernels, thereby allowing custom CNNs to be trained for a specific application (image content and blur range).
Additionally, we evaluate the behavior and limits of the CNNs with respect to blur direction range and length.
\end{abstract}


\section{Introduction}



Motion blur is a typical image degradation caused by the combination of poor lighting conditions with relative movement of scene objects to the camera (e.g. camera shake and object motion).
Motion blur can be reduced by shortening of exposure time at the expense of higher noise, shallower depth of field, higher requirements on imaging sensor quality, or stronger illumination.
Such approaches are not always possible or cost-effective, and deblurring in post-processing stage poses a viable alternative.

Traditional image deconvolution methods are based on a simplified mathematical model of the imaging process which they try to invert, usually in an iterative optimization process.
Motion blur is easily modeled as a convolution with a suitable blur kernel.
However, deblurring may provide unsatisfactory results even when the precise blur kernel is known (non-blind deconvolution) due to image noise or aspects of the imaging process which are not captured by the convolutional model (e.g. color saturation~\cite{Cho2011}, space-variant blur~\cite{whyte2010}, structured noise~\cite{Eigen2013}, and atmospheric effects).
In traffic surveillance systems, especially saturation caused by reflections and light scatter in atmosphere prove challenging for traditional methods.
The reconstruction process becomes even harder in the blind setting when the blur kernel is not known and it has to be estimated from the image.

This paper focuses on an alternative approach to blind deconvolution originally proposed by Hradiš \textit{et al.}~\cite{Hradis2015} which relies on convolutional neural networks (CNN) trained on a large set of artificially blurred images to directly deblur images.
We study this direct deblurring approach in the context of license plate (LP) recognition on blurred images from a real traffic surveillance system.
Similarly to text documents~\cite{Hradis2015}, license plates contain relatively simple and repeated shapes which provide a good data prior for the reconstruction.
Moreover, static placement of the cameras allows us to train custom CNNs for their expected range of motion blur sizes and directions.

Studying the problem of LP recognition provides deeper insights into the properties and capabilities of direct CNN deblurring and, at the same time, it presents valuable information relevant for its commercial applications in traffic surveillance.
Specifically, we show that deblurring CNNs provide superior accuracy of a consequent OCR compared to a state of the art blind deconvolution methods. 
Thereby, we for the first time demostrate state-of-the-art quantitative results of deblurring CNNs on real data.


\begin{figure}
	\centering
	\includegraphics[width=1.\linewidth]{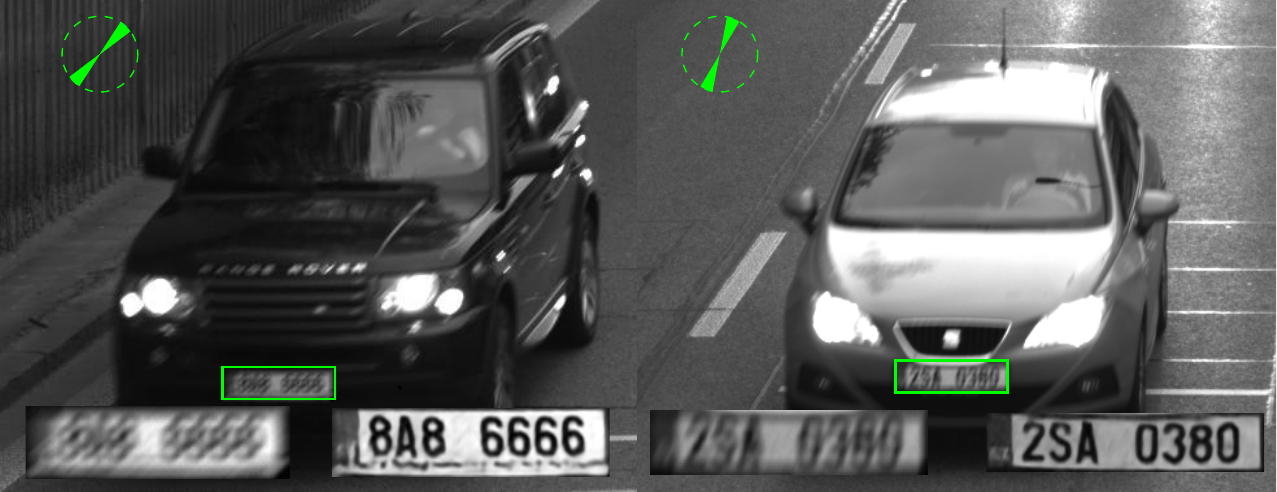}
	\caption{
		License plates from a surveillance system and theirs reconstructions using direct CNN deblurring.
	}
	\label{fig:cnn_car_lp}
\end{figure}



\section{Related Work}
\label{sec:related-work}

The task of blind deconvolution is to find a latent sharp image $x$ by reversing an imaging process which can be in its simplified form defined as
\begin{equation} \label{eq1}
y = x \ast g + n,
\end{equation}
where $g$ is a blur kernel representing degradation due to motion and lens imperfections, and $n$ is additive random noise.

Modern blind deconvolution methods select a suitable data prior, which is transformed into a simple regularizer in an optimization problem which is solved by alternately estimating the blur kernel $\hat{g}$ and the latent image $\hat{x}$~\cite{Pan2014}.
Some of the existing methods rely on rather ad-hoc sharpening steps which seem to be crucial for them to work \cite{Levin2009, Levin2011}.


Several authors aimed to adapt general blind deconvolution methods to license plates.
Fang \textit{et al.} \cite{Fang2015} proposed a new regularization term based on intensities and gradients.
Song \textit{et al.} \cite{Song2015} fuse L0-regularized deblurring with character recognition.
Hsieh \textit{et al.} \cite{Hsieh2010} skip deconvolution and rather focus on recognition of blurred characters.

CNNs have previously been used to learn denoising~\cite{Jain2009}, structured noise removal~\cite{Eigen2013}, non-blind deconvolution~\cite{Schuler2013,Xu2014}, and superresolution~\cite{Dong2014}.

Schuler \textit{et al.}~\cite{Schuler2014} incorporate a small sharpening CNN into an iterative blind deconvolution method in order to better estimate the blur kernel.
Sun \textit{et al.} \cite{Sun2015} specifically address non-uniform blind deconvolution by locally estimating the distribution over 361 motion kernel candidates using a CNN.
The motion estimates are smoothed into a dense motion field by a Markov random field.


\section{CNN for motion deblurring}
\label{sec:cnn-deblurring}

CNNs have been exploited in many state-of-the-art computer vision tasks in recent years, including object and scene classification~\cite{Krizhevsky2012}, object detection~\cite{Girshick2014}, facial recognition~\cite{Taigman2014} and image restoration~\cite{Hradis2015,Jain2009,Eigen2013,Schuler2013,Xu2014,Dong2014}.
Although CNNs have been well known before, they became more important with higher available computational power and especially with GPGPU.


Classification CNNs, labeling an image with a semantic class, are typically constrained to a fixed-sized input~\cite{Krizhevsky2012} due to their fully connected layers.
However, such CNNs can be adapted to process images of arbitrary size by making all their layers convolutional~-- that is, the fully-connected layers are converted to 1-by-1 convolutions~\cite{Sermanet2014,Shelhammer2015}.
Image restoration networks~\cite{Hradis2015,Jain2009,Eigen2013,Schuler2013,Xu2014,Dong2014} are similar to the fully convolutional networks (FCN) for classification and semantic segmentation except they do not use pooling layers -- their output has the same size as their input except cropped borders.




We chose a 15 layer network CNN-L15  which was the largest and most successful network for text deblurring reported by Hradiš \textit{et al.}~\cite{Hradis2015}. The network consists only of convolutional and ReLU layers and it contains 2.2\,M unique weight parameters, i.e. it occupies approximately 9\,MB in memory. 
See Table~\ref{tab:CNN_architecture} for the exact network architecture.

The motion deblurring CNN represented as $F$ consists of an input data layer $F_{0}$, convolutional layers $F_{\ell}$ with their weights represented as convoutional kernels $W_{\ell}$ and their \mbox{biases} $b_{\ell}$:
\begin{equation}
	\begin{split}
		F_{0}(y) &= y \\
		F_{\ell}(y) &= \max(0, W_{\ell} \ast F_{\ell-1}(y) + b_{\ell} )\\
		F(y) &= W_{L} * F_{L-1}(y)+b_{L}
	\end{split}
	\label{eq:cnn-deblur}
\end{equation}


\subsection{Training CNN for motion deblurring}

\begin{figure}
	\centering
	\includegraphics[width=0.85\linewidth]{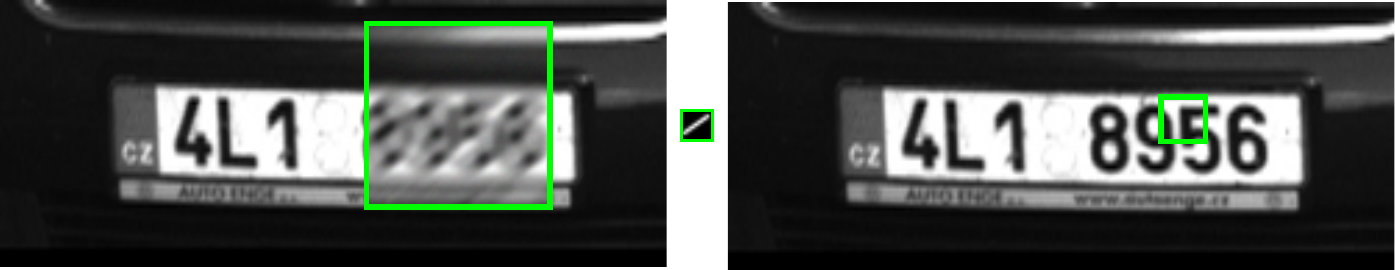}
	\caption{
		An example of training data pair $\{y_i,x_i\}$. Left: network input $y_i$ -- a random blurred crop.	Right: corresponding sharp ground truth $x_i$.
		}
	\label{fig:cnn_train_data}
\end{figure}

CNN-L15 was trained on a dataset of fixed-size patches $D=\{Y,X\}$, where $x_i \in X$ were sharp patches from a surveillance system and $y_i \in Y$ were respective artificially blurred patches. 
The size of the blurred input patches was $66 \times 66$ which corresponds to size of output sharp patches  $16 \times 16$ (see Figure~\ref{fig:cnn_train_data}).
According to~\cite{Hradis2015}, this size provides a reasonable trade off between computational efficiency and diversity of training mini batches.

The network weights were initialized from a uniform distribution with variance $1/n$~\cite{glorot2010}, where $n$ denotes the size of the respective convolutional filter.
The network was trained for 400\,K iterations with mini-batches of 54.
The network took on average 3 days to train on a single NVIDIA GeForce 980 GPU.
Initial learning rate was set to $4 \times 10^{-5}$ and it was reduced three times by a factor of $2$.
The optimized loss function was the sum of squared differences between original and reconstructed patches:
\begin{equation}
\mathcal{L}: \frac{1}{2 |D|} \sum_{i=1}^{|D|} \left\lVert F(y_{i}) - x_{i} \right\rVert_{2}^{2} 
\label{eq:loss}
\end{equation}

\begin{figure*}
	\centering
	\def\svgwidth{0.95\linewidth}
	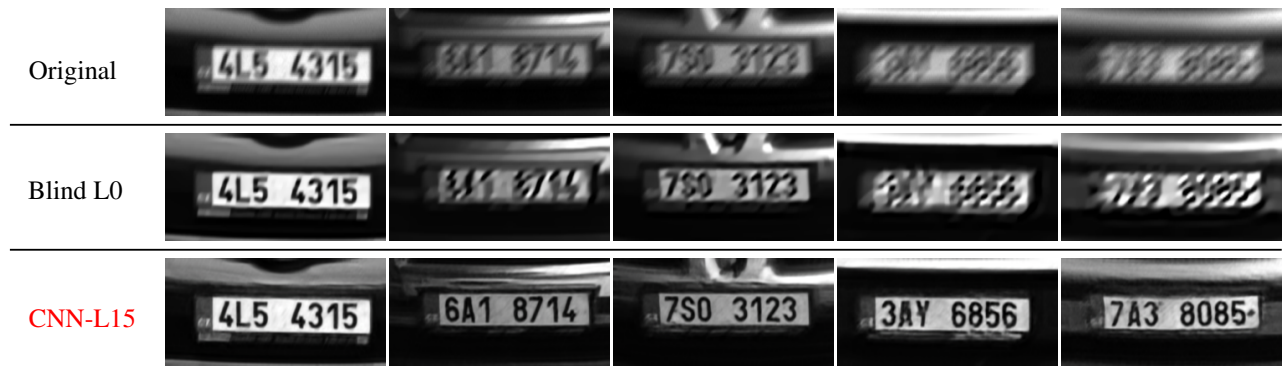
	\caption{ Reconstruction results. First row: naturally blurred images captured by the surveillance system.
	Second row: reconstruction by L0-regularized blind deconvolution~\cite{Pan2014}.
	Third row: results of CNN-L15 trained for blurs with 1--23\,px length and 45 degree wide direction range.
	}
	\label{fig:cnn_reconstruction_results}
\end{figure*}

\section{Results}
\label{sec:experimental-setup}


	

The objective was to explore the ability of CNNs to learn deblurring for different ranges of motion blurs -- specifically, we experiment with different ranges of motion direction and length on artificially blurred images.
The approach was validated on a representative set of naturally blurred images.


\subsection{Data}
\label{sec:datasets}
All images were obtained from two cameras of a traffic surveillance system (examples are shown in Figure~\ref{fig:cnn_car_lp}) from a road with speed limit of 90\,kmph.
The first camera captured cars approaching approximately from direction of 37 to 57 degrees while images from the second one were of cars approaching from about 59 to 79 degrees.

For the validation of the approach, naturally blurred images were collected by increasing exposure time from 6\,ms up to 12\,ms with step of 2\,ms.
This validation dataset contains 721 images almost uniformly distributed between the two cameras and the exposure times.
These images were cropped around the license plate and normalized to $264 \times 128$ px as shown in Figure~\ref{fig:cnn_reconstruction_results}, and they were carefully manually annotated with license plate characters such that OCR accuracy could be evaluated.

A large set of sharp images was gathered for training of CNNs and for fine-grained experiments.
These images were captured with exposure of 2\,ms.
Total 140K sharp images were gathered while approximately 6K images were removed based on ratio of high to low frequencies due to poor contrast or significant blur.
Five randomly cropped patches per image ware taken to create the $(y_i,x_i)$ pairs.
Every patch was artificially blurred with a line motion blur kernel combined with a small depth of field blur kernel (antialiased disk with radius 0-2\,px).
The line kernels were antialiased and their length was sampled with subpixel resolution.
An additive white Gaussian noise with deviation of 0-4 was added to the blurred patches.

The set of 140K images was divided into a training set of 126K images corresponding to 630K patches, and a testing set of 25K patches sampled form 14K images.
The training set was used in all experiments while the test set was used only for the experiments presented in section \nameref{sec:BMresults}.
The validation dataset of the naturally blurred images is available in the supplementary material.
The large set of sharp images is available upon request from the authors.

\begin{table*}[t]
\centering%
	\caption{CNN-L15 architecture -- filter size and number of channels for each layer. }
	\setlength{\tabcolsep}{5pt}
		\begin{tabular}{c|ccccccccccccccc}
		\toprule 
		Layer                  & 1 & 2 & 3 & 4 & 5 & 6 & 7 & 8 & 9 & 10 & 11 & 12 & 13 & 14 & 15 \\
		\midrule
		Filter size & $19{\times}19$ & $1{\times}1$ & $1{\times}1$ & $1{\times}1$ & $1{\times}1$ & $3{\times}3$ & $1{\times}1$ & $5{\times}5$ & $5{\times}5$ & $3{\times}3$ & $5{\times}5$ & $5{\times}5$ & $1{\times}1$ & $7{\times}7$ & $7{\times}7$ \\
	Channel count & 128 & 320 & 320 & 320 & 128 & 128 & 512 & 128 & 128 & 128 & 128 & 128 & 256 & 64 & 3 \\
\bottomrule
\end{tabular}
\label{tab:CNN_architecture}

\end{table*}

\begin{figure*}
	\centering
	\begin{subfigure}[b]{0.33\linewidth}
	\centering
		\includegraphics[width=0.98\linewidth]{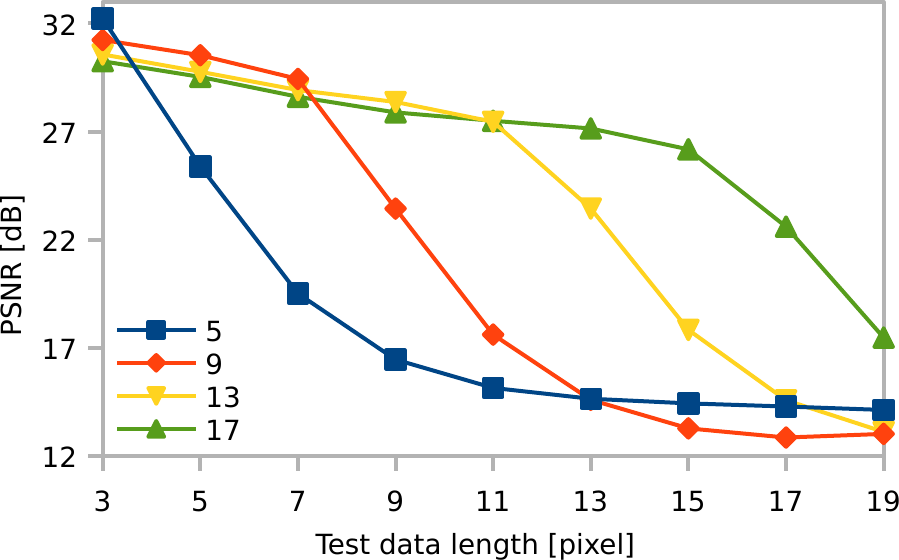}
		\caption{}
		\label{fig:PSNR-a}
	\end{subfigure}%
	\begin{subfigure}[b]{0.33\linewidth}
		\centering
		\includegraphics[width=0.98\linewidth]{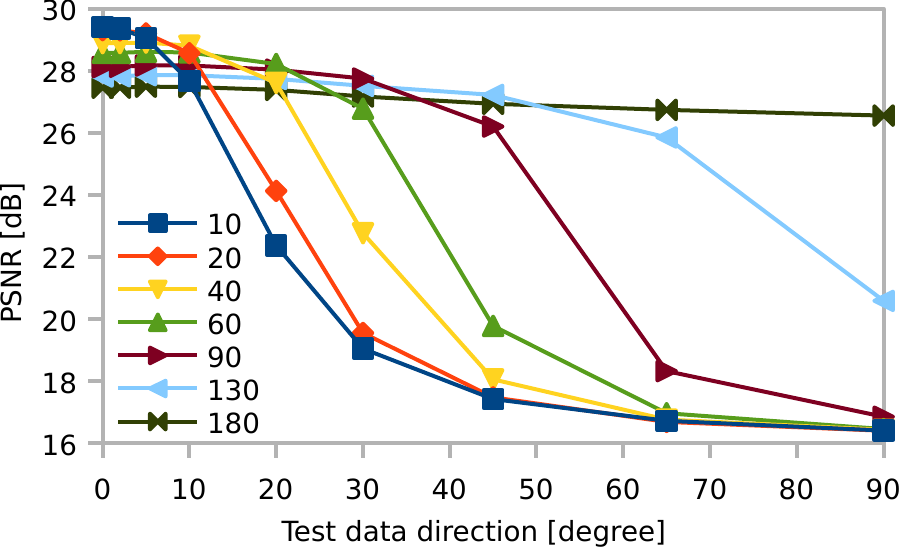}
		\caption{}
		\label{fig:PSNR-b}
	\end{subfigure}%
	\begin{subfigure}[b]{0.33\linewidth}
		\centering
		\includegraphics[width=0.98\linewidth]{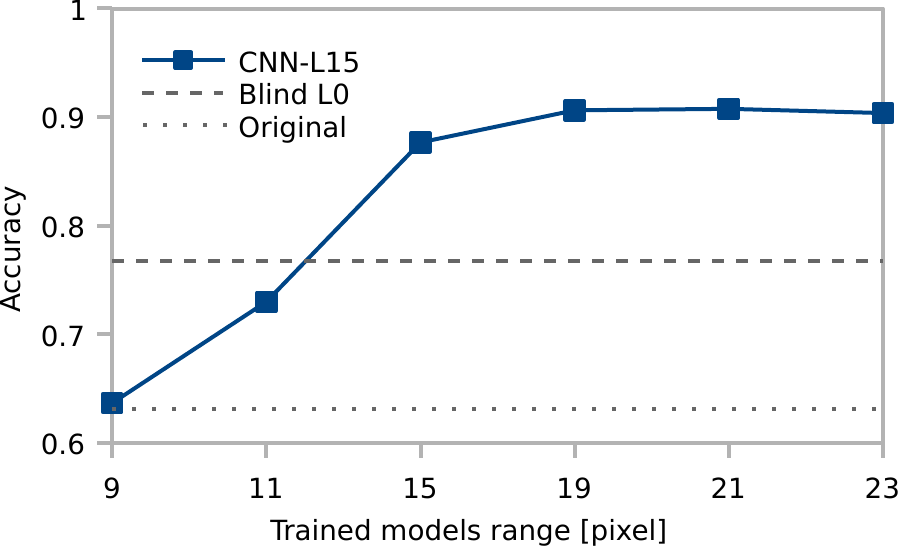}
		\caption{}
		\label{fig:PSNR-c}
	\end{subfigure}%
	\caption{
    (\subref{fig:PSNR-a}) results of CNN-L15 trained for specific length range on different blur lengths.
    (\subref{fig:PSNR-b}) results of CNN-L15 trained for different direction ranges on different deviations from the central direction.    
    (\subref{fig:PSNR-c}) OCR accuracy of 6 CNN-L15 models trained for different length ranges evaluated on the natural dataset compared with L0-regularized blind deconvolution~\cite{Pan2014}.
	}
	\label{fig:PSNR}
\end{figure*}

\subsection{Motion blur range}
\label{sec:BMresults}

Two experiments were performed to asses the behavior of deblurring CNN with respect to motion blur length and with respect to the range of blur direction which it has handle. 
These experiments were performed on the artificially blurred images and the reconstruction image quality was measured as the peak signal to noise ratio (PSNR).

Four networks were trained for different blur lengths 0--5, 0--9, 0--13 and 0--17 pixels, while the direction was uniformly sampled in range of 20 degrees.
Figure~\ref{fig:PSNR-a} shows results of these networks for different blur lengths.
The results show that networks trained for shorter ranges of blur perform better inside these ranges, but their results degrade rapidly outside the range.
The results start to degrade already at the border of the respective ranges probably due to the fact that no larger blurs are represented in the respective training sets.
The reconstruction quality consistently decreases linearly for longer blur kernels.

Second experiment shown in Figure~\ref{fig:PSNR-b} assess performance of the networks for different blur direction ranges. 
Seven networks were trained for 7 direction ranges 10, 20, 40, 60, 90, 130, 180 degrees wide. 
Note that the blur kernels are symmetric and consequently the largest range of 180 degree covers all possible directions.
The blur lengths were all uniformly sampled from 0-13\,px.
The observed results show similar trends as in the experiment with different blur lengths -- networks trained for tighter direction ranges perform better inside these ranges, but their performance degrades rapidly outside the respective direction ranges.

\subsection{Real image deblurring}

We trained six networks to be evaluated the validation dataset of naturally blurred images.
These networks were all trained on blur kernels covering both camera -- the range of blur directions was 50 degrees wide. 
The networks were trained for blur lengths 0--9, 0--11, 0--15, 0--19, 0--21, and 0--23 pixels).

We selected the L0-regularized blind deconvolution method by Pan \textit{et al.}~\cite{Pan2014} as a representative of traditional blind deblurring methods to serve as a baseline. 
This method is specifically optimized for images containing text and it should be suitable for the license plate images.
We selected optimal parameters of the baseline method using grid search directly on the validation dataset. 

Figure~\ref{fig:PSNR-c} shows results on the naturally blurred images as character accuracy of an Optical Character Recognition sytem.
The OCR system we used is optimized for license plates and is used in commercial traffic surveillance systems. 
The networks trained for shorter blur perform poorly as the dataset contains blurs up to 19\,px long. 
Networks trained for sufficiently long blurs significantly outperform the baseline deconvolution methods of Pan \textit{et al.}~\cite{Pan2014}.
The improvement is from character error of 23\% down to 9\% which corresponds to relative improvement by a factor of $2.5$.
It is worth to emphasize that the OCR accuracy does not decrease for models trained for long blur kernels.
Figure~\ref{fig:cnn_reconstruction_results} presents the original images, reconstructed license plates by L0 regularized blind deconvolution and the CNN-L15 reconstructions.

One $264 \times 128$ naturally blurred image takes approx 0.5\,s on NVIDIA GeForce GTX 680.
This is sufficient but only semi-optimal speed due to approach the CNN was utilized.

\section{Conclusion}
\label{sec:conclusion}
We have investigated the ability of CNNs to directly reconstruct motion blurred images of license plates captured by a surveillance system.
The CNNs proved to be effective for the naturally blurred images even though they were trained only on images which were blurred artificially with a simple line kernel. 
The deblurring CNN provided superior accuracy of a consequent OCR compared to a state of the art L0-regularized blind deconvolution tuned for text images~\cite{Pan2014}.
These results show for the first time that CNNs provide quantitatively better deblurring quality compared to state-of-the-art traditional blind methods in a practical application.
The deblurring CNNs are well suited for embedded applications due to the flexibility of the approach, its relatively low computational power requirements, its robustness, and the absence of any tunable parameters.
We suggest the deblurring CNNs to be considered mature and ready to deploy.

The experiments showed that quality of reconstructed images can be improved by customizing the CNNs for a specific range of blurs. 
However, the improvement is only modest in the target application, and general networks trained for a wide range of blurs still provide high-quality results.
The reconstruction quality declines linearly (in PSNR) with the increasing length of the blur kernels which makes it easy to predict achievable reconstruction quality for larger blurs. 
Although the networks are able to reconstruct real images which suggests that the kernels used for training do not have to match the shape of kernels in a real application too closely, the reconstruction quality degrades quite sharply for blurs that are outside the range of blurs the network was trained for.

In the future we would like to train the deblurring CNNs jointly with an OCR network and we would like to incorporate additional sources of image degradation (e.g. atmospheric effects).
An interesting possibility is to reconstruct images directly from JPEG 2000 wavelet coefficients.


\section{Acknowledgement}
This research is supported by the ARTEMIS joint undertaking under grant agreement no 621439 (ALMARVI).
This work was supported by The Ministry of Education, Youth and Sports from the Large Infrastructures for Research, Experimental Development and Innovations project "IT4Innovations National Supercomputing Center – LM\-2015070".

\bibliographystyle{IEEEbib}
\bibliography{./sources}

\end{document}

%% file: figures/results_ocr_images_labeled.pdf_tex
\begingroup%
  \makeatletter%
  \providecommand\color[2][]{%
    \errmessage{(Inkscape) Color is used for the text in Inkscape, but the package 'color.sty' is not loaded}%
    \renewcommand\color[2][]{}%
  }%
  \providecommand\transparent[1]{%
    \errmessage{(Inkscape) Transparency is used (non-zero) for the text in Inkscape, but the package 'transparent.sty' is not loaded}%
    \renewcommand\transparent[1]{}%
  }%
  \providecommand\rotatebox[2]{#2}%
  \ifx\svgwidth\undefined%
    \setlength{\unitlength}{554.84176636bp}%
    \ifx\svgscale\undefined%
      \relax%
    \else%
      \setlength{\unitlength}{\unitlength * \real{\svgscale}}%
    \fi%
  \else%
    \setlength{\unitlength}{\svgwidth}%
  \fi%
  \global\let\svgwidth\undefined%
  \global\let\svgscale\undefined%
  \makeatother%
  \begin{picture}(1,0.28069698)%
    \put(0,0){\includegraphics[width=\unitlength]{results_ocr_images_labeled.pdf}}%
    \put(0.0149343,0.22559872){\color[rgb]{0,0,0}\makebox(0,0)[lb]{\smash{Original}}}%
    \put(0.01430912,0.13092679){\color[rgb]{0,0,0}\makebox(0,0)[lb]{\smash{Blind L0}}}%
    \put(0.0149343,0.03042893){\color[rgb]{1,0,0}\makebox(0,0)[lb]{\smash{CNN-L15}}}%
  \end{picture}%
\endgroup%